\pdfoutput=1

\documentclass[11pt]{article}

\usepackage[]{ACL2023}

\usepackage{times}
\usepackage{latexsym}

\usepackage[T1]{fontenc}

\usepackage[utf8]{inputenc}

\usepackage{microtype}

\usepackage{inconsolata}

\usepackage[utf8]{inputenc}
\usepackage{booktabs}
\usepackage{graphicx}
\usepackage{CJK}
\usepackage{multirow}
\usepackage{color}
\usepackage{verbatim}
\usepackage{url}
\usepackage{enumitem}
\usepackage{amsmath}
\usepackage{diagbox}
\usepackage{flushend,cuted}
\usepackage{bbm}
\usepackage{xcolor}
\usepackage{xspace}
\usepackage{algorithm}
\usepackage{algpseudocode}
\usepackage{verbatim}
\usepackage{colortbl}
\usepackage{bbding}

\usepackage{listings}
\usepackage{ulem}

\newcommand{\method}{PLAC\xspace}
\newcommand{\metric}{\textit{knowledge margin}\xspace}
\newcommand{\knn}{$k$NN\xspace}
\newcommand{\knnmt}{$k$NN-MT\xspace}
\newcommand{\atat}{\resizebox{12pt}{\totalheight}{@@}\xspace}

\newcommand\kt[1]{\begin{CJK*}{UTF8}{gkai} #1 \end{CJK*}}

\title{What Knowledge Is Needed? Towards Explainable Memory for $k$NN-MT Domain Adaptation}

\author{
    Wenhao Zhu$^{1,2}$, 
    Shujian Huang$^{1,2}$, 
    Yunzhe Lv$^{1,2}$, 
    Xin Zheng$^{1,2}$, 
    Jiajun Chen$^{1,2}$ \\
    $^{1}$ National Key Laboratory for Novel Software Technology, Nanjing University, China \\ 
    $^{2}$ Collaborative Innovation Center of Novel Software Technology and Industrialization \\
    {\{zhuwh, lvyz, zhengxin\}@smail.nju.edu.cn, \{huangsj, chenjj\}@nju.edu.cn} \\
}

\begin{document}
\maketitle

\begin{abstract}
$k$NN-MT presents a new paradigm for domain adaptation by building an external datastore, which usually saves all target language token occurrences in the parallel corpus.
As a result, the constructed datastore is usually large and possibly redundant.
In this paper, we investigate the interpretability issue of this approach: what knowledge does the NMT model need?
We propose the notion of local correctness (LAC) as a new angle, which describes the potential translation correctness for a single entry and for a given neighborhood.
Empirical study shows that our investigation successfully finds the conditions where the NMT model could easily fail and need related knowledge.
Experiments on six diverse target domains and two language-pairs show that pruning according to local correctness brings a light and more explainable memory for $k$NN-MT domain adaptation.
\end{abstract}


\section{Introduction}
Domain adaptation in neural machine translation (NMT) aims at adapting pre-trained NMT models to a target domain \cite{chu2017empirical, thompson2019overcoming, hu2019domain,zhao2020knowledge, zheng2021adaptive}.
Fine-tuning \cite{luong2015stanford} has been the de facto standard 
for adaptation. 
However, fine-tuning suffers from the catastrophic forgetting problem \cite{mccloskey1989catastrophic, french1999catastrophic}.

Recently, \citet{khandelwal2021nearest} propose $k$NN-MT, showing a new paradigm for domain adaptation. 
$k$NN-MT first explicitly extracts translation knowledge in the target domain training data into a \textit{key}-\textit{value} datastore with a pre-trained NMT model.
For each datastore entry, the key is a continuous representation and the value is a symbolic token.
The datastore is then used to assist the NMT model during translation. 
The $k$NN-MT framework circumvents the necessity to disturb the parameters of the pre-trained NMT model and enables quick adaptation by switching datastores.


$k$NN-MT incorporates the symbolic datastore to assist the neural model \cite{khandelwal2021nearest, zheng2021adaptive, jiang2021learning}.
However, the datastore usually stores all the target tokens in the parallel data, without considering the capability of the neural model. 
As a result, the datastore is usually huge in size and possibly redundant.

To understand the relationship between the datastore and the NMT model, this paper conducts investigations on the interpretability issue: \textit{what knowledge does the NMT model need?} 
Intuitively, the pre-trained NMT model only needs knowledge that remedies its weaknesses.
Thus, we propose to explore this issue from the point of \textit{local correctness} (Section \ref{sec:analysis}). 
Our local correctness includes two aspects, the correctness of translating a given entry (\textit{entry correctness}) and, more importantly, the correctness of performing translation in a given neighborhood in the representation space (\textit{neighborhood correctness}). 

For the entry correctness, we check whether the NMT could make correct translation for the entry itself and accordingly split the datastore entries into two categories, namely \textit{known} and \textit{unknown}.
Based on entry correctness, we examine neighborhood correctness to more comprehensively evaluate the NMT model's underlying capability.
Specifically, we propose a \metric metric to evaluate the maximum size of the neighborhood where the NMT could make correct translation.
Intuitively, the NMT model may fail when the knowledge margin is small.

To verify our interpretation, we devise a datastore pruning algorithm \method (\textbf{P}runing with \textbf{L}oc\textbf{A}l \textbf{C}orrectness), which simply removes entries with a higher knowledge margin value (Section \ref{sec:method}). 
These entries are less useful for adaptation, because the NMT model translates well in their neighborhood.

We conduct experiments on six diverse target domains in two language pairs (Section \ref{sec:results}).
Compared with existing pruning baselines \cite{martins2022efficient, wang2022efficient}, \method prunes more entries (up to 45\%) in four OPUS domains' datastore without hurting translation performance.
Through ablation study, we reveal that simply relying on entry correctness is not enough, showing that the novel metric knowledge margin for the neighborhood correctness could be the key to build a light and more explainable memory for $k$NN-MT domain adaptation.



\section{Background}
For NMT domain adaptation, $k$NN-MT constructs a datastore $\mathcal{D}$ based on the given target domain bilingual corpus $\mathcal{C}$ and use it to provide helpful target domain translation knowledge for the pre-trained NMT model $\mathcal{M}$.
In this section, we briefly introduce $k$NN-MT and its advanced variant, adaptive $k$NN-MT \cite{zheng2021adaptive}.

\subsection{Building a Domain Specific Datastore}
Given target domain bilingual corpus $\mathcal{C}$, all translation pairs in $\mathcal{C}$ are fed into the frozen pre-trained NMT model for decoding with teacher-forcing \cite{williams1989learning}.
At decoding time step $t$, the hidden state from the last decoder layer $h(\mathbf{x}, \mathbf{y}_{<t})$ is taken as key and the $t$-th target token $y_t$ is taken as value, resulting in a key-value pair.
For the entire corpus, the datastore $\mathcal{D}$ is consisted of key-value pairs:
\begin{equation}
    \mathcal{D} = \{(h(\mathbf{x}, \mathbf{y}_{<t}), y_t) ~|~ \forall y_t \in \mathbf{y}, (\mathbf{x}, \mathbf{y})\in \mathcal{C}\},
\end{equation}
where $\mathbf{y}_{<t}$ denotes previous tokens in the sequence $\mathbf{y}$.
Each entry in the datastore explicitly memorizes the following translation knowledge: generating the value token at the  decoder hidden state key. And the datastore covers all target language token occurrences. 

\subsection{Translating with the Datastore}
During inference, given a source language sentence $\mathbf{x}$, 
$k$NN-MT simultaneously leverages $\mathcal{M}$ and $\mathcal{D}$ to generate target language translation $\mathbf{y}\textrm{=}\{y_1, y_2, \cdots, y_{|\mathbf{y}|}\}$.
More specifically, at decoding time step $t$, $k$NN-MT queries the datastore with the decoder hidden state $h(\mathbf{x}, \mathbf{y}_{<t})$ generated by $\mathcal{M}$.
The $k$ nearest neighbors of the query $\mathcal{N}_k=\{(h^j, y^j)\}_1^k$ are retrieved, which are $k$ entries with keys closest to the query according to squared-$L^2$ distance, $d$. These retrieved knowledge are converted into a distribution over the vocabulary:
\begin{align}
p_{\textrm{$k$NN}}&(y_t |\mathbf{x}, \mathbf{y}_{<t}) \propto \\ 
&\sum_{(h^j, y^j)\in \mathcal{N}_k}\mathbbm{1}_{y_t=y^j}\exp(\frac{-d(h^j,h(\mathbf{x},\mathbf{y}_{<t}))}{T}), \nonumber 
\end{align}
where $T$ is the temperature.
Then, $k$NN-MT interpolates $p_{\textrm{$k$NN}}$ with the pre-trained NMT model's output distribution as the final translation distribution:
\begin{equation}
\begin{split}
p(y_t|\mathbf{x}, \mathbf{y}_{<t}) &= \lambda \ p_{\textrm{$k$NN}}(y_t|\mathbf{x}, \mathbf{y}_{<t}) \\
& + (1-\lambda) \ p_{\textrm{NMT}} (y_t|\mathbf{x}, \mathbf{y}_{<t})
\end{split}
\end{equation}
The complete translation $\mathbf{y}$ can be generated by beam search.

\subsection{Adaptive $k$NN-MT}
For vanilla $k$NN-MT, the selection of hyper-parameters, such as $k$ or $\lambda$, highly affect the final translation performance, which is less stable across languages or domains.
Adaptive $k$NN-MT uses a lightweight meta-$k$ neural network to dynamically determine the usage of retrieved entries, which avoids the tuning of hyper-parameters and achieves a more stable performance \cite{zheng2021adaptive}.

\section{What Knowledge Does the NMT Model Need?}
\label{sec:analysis}
Although less accurate, the pre-trained NMT model could perform translation without the datastore. 
This fact suggests that the NMT model knows some bilingual knowledge of the target domain.
However, the construction of datastore dismisses this point and results in a huge amount of entries being stored.

Intuitively, the pre-trained NMT model only needs knowledge that remedies its weaknesses.
To find out these weaknesses and build more explainable memory, we start from investigating entry correctness.
Based on this basic concept, we further study neighborhood correctness and find that it precisely reflects the NMT model's strengths and weaknesses. 


\subsection{Known v.s. Unknown for Entry Correctness}
\label{sec:not_new}
The capability of the NMT model in target domain is difficult to describe directly.
However, as the datastore consists of entries constructed on training set, it is easier to check whether the NMT model could make correct translation for them.

This can be efficiently accomplished by an extra evaluation during the teacher-forcing decoding. 
More specifically, at each time step $t$ of the teacher-forcing process, we not only record the hidden states $h(\mathbf{x}, \mathbf{y}_{<t})$ and the correct target token $y_t$, but also evaluate the prediction of the NMT model $y'_t$, which is the target token with the highest probability $p_{\textrm{NMT}} (y'_t|\mathbf{x}, \mathbf{y}_{<t})$ .
Then we call an entry as a \textit{known} entry if the NMT model could predict it correctly; and \textit{unknown}, otherwise (Equation~\ref{equ:known}).
\begin{equation}
(h(\mathbf{x}, \mathbf{y}_{<t}),y_t) \text{ is }
\begin{cases}
\textit{known}, &\text{if $y'_t=y_t$} \\
\textit{unknown},&\text{o.w.}  \\
\end{cases}\label{equ:known}
\end{equation}

Obviously, the \textit{unknown} entries in the datastore are important, because these are the points where the NMT model tends to make a mistake.

\subsection{The Knowledge Margin Metric for Neighborhood Correctness}
\label{sec:km}
However, entry correctness alone could not fully reveal the NMT model's weaknesses.
Because for \textit{known} entries, the NMT model may still fail during inference where the context could be similar but different.
Considering that the contextualized representations of similar context stay close in the representation space \cite{peters2018deep},
 we propose to investigate the NMT model's translation performance in a neighborhood.



We propose a metric called knowledge margin, denoted as $km$, to measure the neighborhood correctness.
Given an entry $(h, y)$, its neighborhood is defined by its $k$ nearest neighbors\footnote{In our implementation, we do not consider the given entry itself as its neighbor.} in the datastore $\mathcal{N}_k(h)=\{(h^j, y^j)\}_1^k$. 
The knowledge margin of the entry, i.e. $km(h)$, is defined as:

\begin{equation}   
\arg\max_k {(h^j, y^j)\textrm{ is \textit{known}, } \forall (h^j, y^j) \in \mathcal{N}_k (h) }.\
\end{equation}\label{equ:margin}
Intuitively, $km$ is the maximum size of the neighborhood of the entry $h$ where the NMT could make correct translation. 
If considering at most $\bar{k}$ nearest neighbors of $h$, its knowledge margin will be a number between 0 and $\bar{k}$. 

Please note that the definition of knowledge margin applies for any point in the representation space, because for each point (e.g. an actual query $q$ during inference), its neighborhood $\mathcal{N}_k(q)$ could be defined by querying the datastore. This extension allows the investigation of the NMT model at any given point in the representation space.

\subsection{Empirical Analysis}
We now present an empirical analysis of the relationship between the NMT model and the datastore, and reveal the NMT model's weaknesses.

\paragraph{Settings}
We follow \citet{zheng2021adaptive} and consider four domains in German-English OPUS dataset \cite{tiedemann2012parallel} as target domains\footnote{Detailed description of OPUS domains can be found in Appendix \ref{sec:description}.}.
Table \ref{tab:opus} lists statistics of four domains\footnote{We use the dataset re-split by \citet{aharoni-goldberg2020unsupervised}, which removes the overlap between training, development and test sets.}.
For pre-trained NMT model, we use the winner model of WMT'19 German-English news translation task \footnote{\url{https://github.com/pytorch/fairseq/tree/main/examples/wmt19}}~\cite{ng2019facebook}.
The datastore for each domain is constructed on the corresponding training set with the pre-trained NMT model.

\begin{table}[htbp]
    \footnotesize
    \centering
    \scalebox{0.9}{
        \begin{tabular}{l|cccc}
        \toprule
                    & OPUS-   & OPUS-     & OPUS-      & OPUS- \\
                    & Medical & Law       &  IT        & Koran \\
        \midrule 
        Train       & 248,099 & 467,309 & 222,927 & 17,982 \\
        Dev         & 2,000   & 2,000   & 2,000   & 2,000 \\
        Test        & 2,000   & 2,000   & 2,000   & 2,000 \\
        \bottomrule
        \end{tabular}
    }
    \caption{Number of sentences of the OPUS dataset. ``Train'', ``Dev'', ``Test'' denote training, development, test set, respectively.}
    \label{tab:opus}
\end{table}

\paragraph{Entry Correctness}
We collect statistics about the two categories of entries and report results in Table \ref{tab:train_acc}.
The results show that 56\%$\sim$73\% (averaging 66.7\%)
of datastore entries are \textit{known} by the pre-trained NMT model. 
This high ratio strongly indicates that a large amount of datastore entries may be redundant. 

\begin{table}[htbp]
    \footnotesize
    \centering
    \scalebox{0.9}{
        \begin{tabular}{p{1.8cm}<{\centering}|p{1.15cm}<{\centering}p{1.15cm}<{\centering}p{1.15cm}<{\centering}p{1.15cm}<{\centering}}
        \toprule
                   &  OPUS-   & OPUS- & OPUS- & OPUS- \\
                   &  Medical & Law & IT & Koran \\
        \midrule
        \textit{known}   &  5,070,607 & 14,803,149 & 2,514,757 & 294,094 \\
        \textit{unknown} &  1,844,966 & 4,287,906  & 1,093,974 & 230,677 \\
        $|\mathcal{D}|$   &  6,915,573 & 19,091,055 & 3,608,731 & 524,771 \\
        \midrule
        \textit{known ratio}             &  73.32\%   & 66.74\%    & 69.69\%   & 56.04\% \\
        \bottomrule
        \end{tabular}
    }
    \caption{The statistics of the \textit{known} and \textit{unknown} entries for the pre-trained NMT model on four OPUS domains' training set. The number of entries and the ratio of \textit{known} entries are listed.}
    \label{tab:train_acc}
\end{table}

\begin{figure*}[ht]
    \centering
    \includegraphics[width=\textwidth]{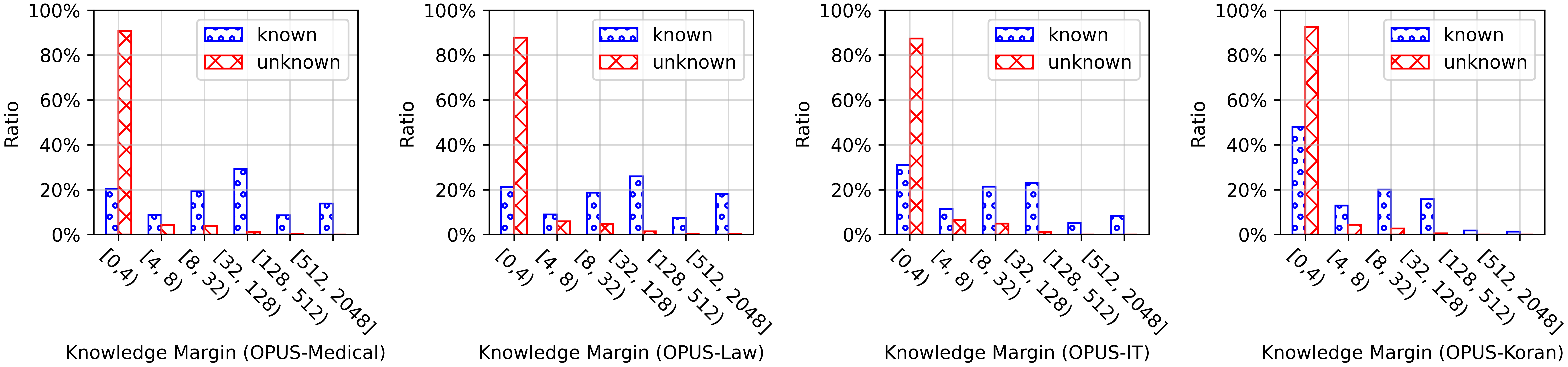}
    \caption{The ratio distribution on different knowledge margin for \textit{known} and \textit{unknown} entries on four OPUS domains.
    }
    \label{fig:maxmargin}
\end{figure*}


\paragraph{Neighborhood Correctness}
We measure neighborhood correctness of each datastore entries and plot the distribution of knowledge margin for known and unknown entries in Figure \ref{fig:maxmargin} ($\bar{k}=2048$).
The distributions on four OPUS domains show the same trends. 
Most \textit{unknown} entries has a very low knowledge margin, e.g., around 90\% of \textit{unknown} entries have a margin value between 0 and 4.
In contrast, the distribution for \textit{known} entries is more diverse. The results indicate that the neighborhood correctness is consistent with the entry correctness, but may provide more information for known entries.




To verify the relation between knowledge margin and NMT model's translation ability, we conduct experiments on the development set for each domain, where translation context are unseen.
For each token $y_t$ in the dev set, we perform teacher-forcing until time step $t-1$ and query the datastore for the neighborhood at time step $t$.
We evaluate the knowledge margin of the query and the prediction accuracy of the NMT model.

\begin{figure}[ht]
    \centering
    \includegraphics[width=0.5\textwidth]{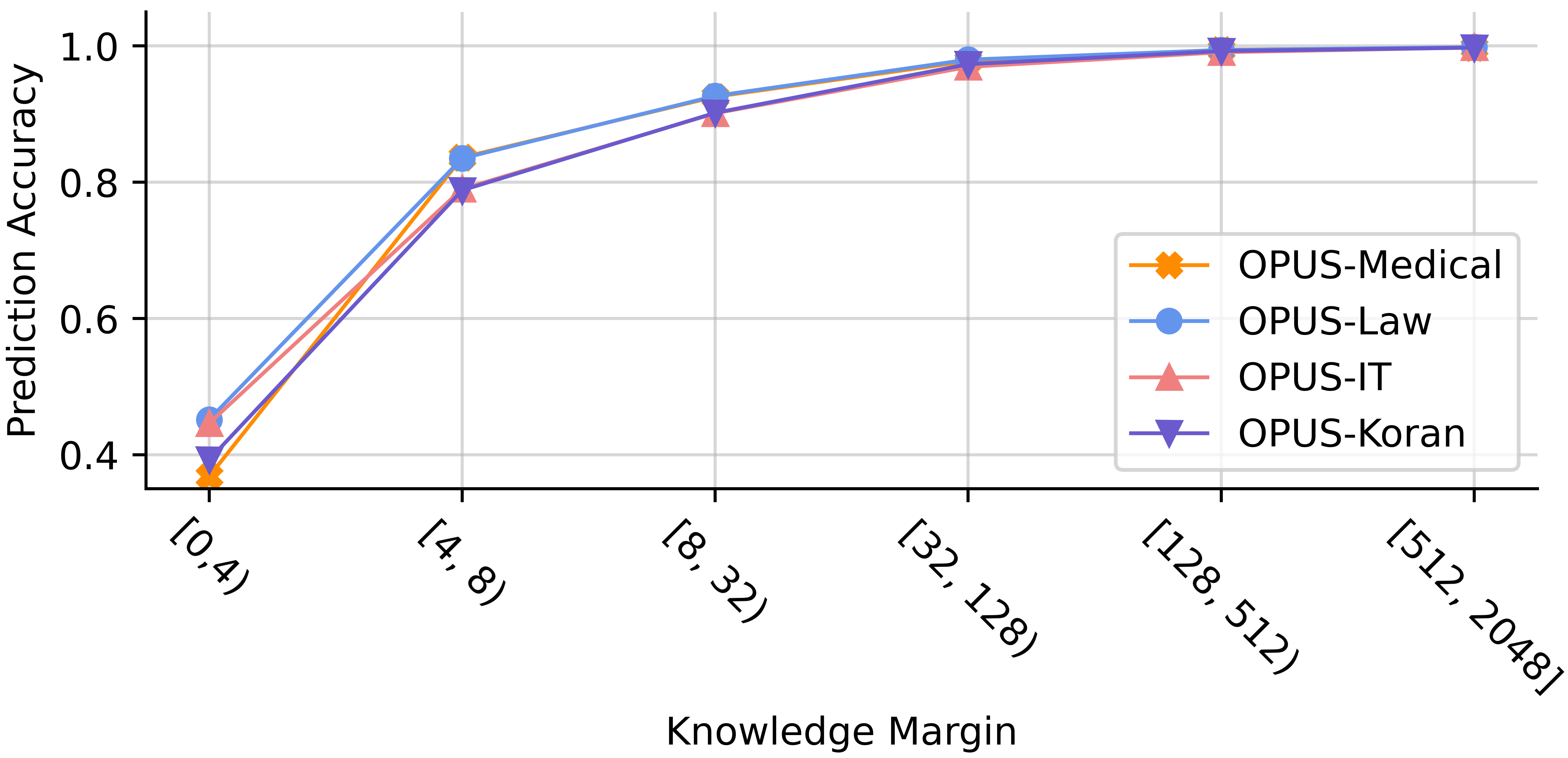}
    \caption{The NMT model's prediction accuracy at positions with different margin values in OPUS domains' unseen development set.}
    \label{fig:tgt_knn_pred}
\end{figure}

Figure \ref{fig:tgt_knn_pred} shows the results. 
For tokens with higher margins, e.g. $km \geq 32$, the prediction accuracy of the NMT model is higher than 95\%. 
In contrast, for tokens with lower margins, e.g. $km<4$, the accuracy is lower than 50\%.
This is a strong evidence that the NMT model could easily fail when knowledge margin is small.

In Table \ref{tab:valid_case}, we also show a translation example for such a condition, where knowledge margin of the current query is 0 and the NMT model fails to generate the last subword of ``Cyanokit''\footnote{``Cyanokit'' is a drug name, which contains the active substance hydroxocobalamin (vitamin B12).}.




\begin{table*}[ht]
    \footnotesize
    \centering
    \scalebox{1}{
        \begin{tabular}{p{0.2cm}<{\centering}p{1.2cm}<{\centering}p{6.1cm}p{4.4cm}p{2.3cm}<{\centering}}
        \toprule
        \multicolumn{5}{l}{\textbf{Source sentence ($\mathbf{x}$):} Wie ist Cy\atat an\atat ok\atat it anzu\atat wenden ?} \\
        \multicolumn{5}{l}{\textbf{Previous translation ($\mathbf{y}_{<t}$):} How to use Cy\atat an\atat ok\atat} \\
        \midrule
        \textbf{No.} & \textbf{Type} & \textbf{Retrieved Keys: source ($\mathbf{x}$)} &  \textbf{Retrieved Keys: target ($\mathbf{y}_{<t}$)} & \textbf{Retrieved Values} \\
        \midrule
        1 & \textit{unknown} & Wie wird Cy\atat an\atat ok\atat it ange\atat wendet ? & How is Cy\atat an\atat ok\atat & it  \\
        2 & \textit{unknown} & Sie erhalten Cy\atat an\atat ok\atat it als In\atat fusion in eine V\atat ene . & You will have Cy\atat an\atat ok\atat & it \\
        3 & \textit{unknown} & Wo\atat f\"ur wird Cy\atat an\atat ok\atat it ange\atat wendet ? & What is Cy\atat an\atat ok\atat & it \\
        4 & \textit{unknown} & Wel\atat ches Risiko ist mit Cy\atat an\atat ok\atat it verbundenn ? & What is the risk associated with Cy\atat an\atat ok\atat & it  \\
        5 & \textit{unknown} & Die folgenden Neben\atat wirkungen wurden in Verbindung mit der Anwendung von Cy\atat an\atat ok\atat it berichtet . & The following un\atat desi\atat rable effects have been reported in association with Cy\atat an\atat ok\atat & it  \\
        6 & \textit{unknown} & Warum wurde Cy\atat an\atat ok\atat it zugelassen ? & Why has Cy\atat an\atat ok\atat & it  \\
        7 & \textit{unknown} & Beson\atat dere Vorsicht bei der Anwendung von Cy\atat an\atat ok\atat it ist erforderlich & Take special care with Cy\atat an\atat ok\atat & it \\
        8 & \textit{unknown} & Wie wirkt Cy\atat an\atat ok\atat it ? & How does Cy\atat an\atat ok\atat & it \\
        \midrule
        \multicolumn{5}{l}{\textbf{NMT's prediction  ($y'_t$):} ite } \\
        \multicolumn{5}{l}{\textbf{Correct target token ($y_t$):} it} \\
        \bottomrule
        \end{tabular}
    }
    \caption{An example where the NMT model fails (sentence are tokenized into subwords). At the current time step, all retrieved entries are \textit{unknown} for the NMT model, so knowledge margin is 0. The prediction of NMT is highly likely to be wrong. With these retrieved entries, the $k$NN-MT could make a correct prediction.}
    \label{tab:valid_case}
\end{table*}

\section{Building Explainable Memory Based on Local Correctness}
\label{sec:method}
Because local correctness are good indicators for translation failures. It could also be interpreted as the importance of datastore entries.
 To verify this interpretation, 
we propose a pruning algorithm, i.e. \textbf{P}runing with \textbf{L}oc\textbf{A}l \textbf{C}orrectness (\textbf{\method}), to cut off entries with a high knowledge margin (Algorithm \ref{alg:pruning}).

There are two steps in the algorithm.
In the first step, each entry $(h, y)$ in the datastore $\mathcal{D}$ is checked for their local correctness. 
If knowledge margin of $(h, y)$ is greater than or equal to the threshold $k_p$, the entry is collected as the pruning candidates \footnote{In practice, there exists a very small amount of \textit{unknown} entries with high knowledge margin. In our implementation, we keep them in the datastore because the NMT model cannot make correct translation about these entries.}.

In the second step, these pruning candidates are randomly selected and get removed from $\mathcal{D}$ until the required pruning ratio is reached.
Since our method does not need to train any additional neural networks, it can be easily implemented.
The pruned datastore can be used in different $k$NN-MT models, such as adaptive $k$NN-MT.
\begin{algorithm}[ht]
\renewcommand{\algorithmicrequire}{\textbf{Input:}}
\renewcommand{\algorithmicensure}{\textbf{Output:}}
\caption{Datastore Pruning by \method}\label{alg:pruning}
\begin{algorithmic}[1]
\footnotesize
\Require datastore $\mathcal{D}$, the \metric threshold $k_p$, the pruning ratio $r$
\Ensure pruned datastore $\mathcal{D}$ \\
$candidates$ $\gets$ $\emptyset$ 
\Comment{step 1: collect}
\For{ each entry $(h, y)$ in $\mathcal{D}$}
    \If {$km(h) \geq k_p$}: 
        \State $candidates \gets candidates \cup (h, y)$
    \EndIf
\EndFor
\Repeat
\Comment{step 2: drop}
\State randomly select entry $(h, y)$ from $candidates$ 
\State remove $(h, y)$ from $\mathcal{D}$
\Until{pruning ratio $r$ is satisfied} \\
\Return $\mathcal{D}$
\end{algorithmic}
\end{algorithm}

\section{Experiment Setup}
This section introduces general experiment setup for evaluating pruning effect. 
More implementation details can be found in Appendix \ref{apx:impl}. 

\subsection{Data and Processing}
We conduct datastore pruning for 6 different domains from 2 language pairs.
Specifically, we take 4 OPUS domains for De-En experiments and 2 UM domains\footnote{We split the original training set into training, development, test set because there is no development set provided in the original dataset and there exists an overlap between original training and test sets. Detailed description of UM domains can be found in Appendix \ref{sec:description}.} for Zh-En experiments~\cite{tian2014corpus}, which are all benchmark dataset for NMT domain adaptation research. 

\begin{table}[htbp]
    \footnotesize
    \centering
    \scalebox{0.9}{
        \begin{tabular}{l|cc}
        \toprule
                    & UM-Law     & UM-Thesis   \\
        \midrule 
        Train       & 216,000 & 296,000 \\
        Dev         & 2,000   & 2,000   \\
        Test        & 2,000   & 2,000   \\
        \bottomrule
        \end{tabular}
    }
    \caption{Detailed statistics of UM dataset. We report the sentence number of each subset. ``Train'', ``Dev'', ``Test'' denote training, development, test set respectively.}
    \label{tab:um}
\end{table}

For preprocessing, we use \textit{moses}\footnote{\url{https://github.com/moses-smt/mosesdecoder}} toolkit to tokenize German and English corpus and \textit{jieba}\footnote{\url{https://github.com/fxsjy/jieba}} to tokenize Chinese corpus.
Byte pair encoding\footnote{\url{https://github.com/rsennrich/subword-nmt}} (BPE) is applied for subword segmentation.


\begin{table*}[tbp]
    \footnotesize
    \centering
    \scalebox{0.9}{
    \begin{tabular}{p{1.9cm}|p{0.6cm}<{\centering}p{0.7cm}<{\centering}p{1.3cm}<{\centering}|p{0.6cm}<{\centering}p{0.7cm}<{\centering}p{1.3cm}<{\centering}|p{0.6cm}<{\centering}p{0.7cm}<{\centering}p{1.3cm}<{\centering}|p{0.6cm}<{\centering}p{0.7cm}<{\centering}p{1.3cm}<{\centering}}
    \toprule
               & \multicolumn{3}{c|}{OPUS-Medical} & \multicolumn{3}{c|}{OPUS-Law} & \multicolumn{3}{c|}{OPUS-IT} & \multicolumn{3}{c}{OPUS-Koran} \\
               & Ratio & BLEU$\uparrow$ &  COMET$\uparrow$ & Ratio & BLEU$\uparrow$ &  COMET$\uparrow$ & Ratio & BLEU$\uparrow$ &  COMET$\uparrow$ & Ratio & BLEU$\uparrow$ &  COMET$\uparrow$ \\
    \midrule
    Base           &  -        & 39.73   & 0.4665 & -     & 45.68 & 0.5761  & -     & 37.94 & 0.3862   & -     & 16.37   & -0.0097 \\
    Finetune          &  -        & 58.09  & 0.5725 & -     & 62.67 & 0.6849 & -     & 49.08 & 0.6343 & -     &  22.40  & 0.0551 \\
    Adaptive $k$NN  &  0\%      & 57.98   & 0.5801 & 0\%   & 63.53 & 0.7033 & 0\%   & 48.39 & 0.5694  & 0\%   & 20.67   & 0.0364 \\
    \midrule
    \textbf{Random} & 45\%               & $\textrm{54.08}^*$    & 0.5677 & 45\%  & $\textrm{58.69}^*$ & $\textrm{0.6690}^*$   & 40\%  & $\textrm{45.54}^*$ & $\hspace{0.15cm}\textrm{0.5314}^*$  & 25\%  & 20.36   & 0.0434 \\
    \textbf{Cluster}           & 45\%      & $\textrm{53.31}^*$  & 0.5689 & 45\%  & $\textrm{58.68}^*$ & $\textrm{0.6779}^*$   & 40\%  & $\textrm{45.80}^*$ &                    0.5788 & 25\%   & $\textrm{20.04}^*$ & 0.0410\\
    \textbf{Merge}          & 45\%         & $\textrm{54.65}^*$  & \hspace{0.15cm}$\textrm{0.5523}^*$ & 45\%  & $\textrm{60.60}^*$ & $\textrm{0.6776}^*$   & 40\%  & $\textrm{45.83}^*$ &  $\hspace{0.15cm}\textrm{0.5334}^*$ & 25\%  & $\textrm{20.25}^*$   & 0.0365 \\
    \textbf{Known} & 45\%                & $\textrm{56.44}^*$    & 0.5691 & 45\%  & $\textrm{61.61}^*$ & $\textrm{0.6885}^*$   & 40\%  & $\textrm{45.93}^*$ & 0.5563 & 25\%  & $\textrm{20.35}^*$  & 0.0338 \\
    \textbf{All Known} & 73\%            & $\textrm{42.73}^*$    & \hspace{0.15cm}$\textrm{0.4926}^*$ & 66\%  & $\textrm{51.90}^*$ & $\textrm{0.6200}^*$   & 69\%  & $\textrm{40.93}^*$ & $\hspace{0.15cm}\textrm{0.4604}^*$ & 56\%  & $\textrm{17.76}^*$  & $\hspace{0.15cm}\textrm{0.0008}^*$ \\
    \midrule
    \textbf{\method} (ours)  & 45\%      & 57.66  & 0.5773              & 45\%  & 63.22              & $\textrm{0.6953}^*$   & 40\%  & 48.22              & 0.5560             & 25\%   & 20.96 & 0.0442 \\
    \bottomrule
    \end{tabular}
    }
    \caption{Pruning Effect on four OPUS domains. ``Ratio'' denotes the pruning ratio. Higher ``BLEU'' and ``COMET'' scores indicates better translation quality. ``*'' means that performance decline is statistically significant ($p<0.05$).}
    \label{tab:opus_test}
\end{table*}
\subsection{Pre-trained NMT Model}
For De-En tasks, we use the winner model of WMT'19 De-En news translation task, which is based on the Transformer architecture \cite{vaswani2017attention}.
For Zh-En tasks, we train a base Transformer model from scratch on CWMT'17 Zh-En Dataset\footnote{\url{http://nlp.nju.edu.cn/cwmt-wmt}} (9 million sentence pairs), since we do not find any publicly available Zh-En pre-trained NMT model on the website.

The pre-trained NMT model is the unadapted general domain model for each language pair, which is the starting point for domain adaptation. For \knn methods, it also serves as the base for building the datastore.

\subsection{Systems for Comparison}
We report the performance of the following systems for reference: the  pre-trained NMT model ({Base}), the pre-trained model finetuned on each target domain ({Finetune}) \cite{luong2015stanford}, adaptive $k$NN-MT with full datastores built for each target domain on their training set (Adaptive \knn) \cite{zheng2021adaptive}. 
Finetuning and Adaptive \knn are two popular alternatives for adaptation.

The following pruning methods are applied to the datastore of Adaptive \knn for comparison: randomly pruning (\textbf{Random}), cluster-based pruning (\textbf{Cluster}) \cite{wang2022efficient}, merging similar entries (\textbf{Merge}) \cite{martins2022efficient}, randomly pruning \textit{known} entries (\textbf{Known}), pruning all \textit{known} entries (\textbf{All Known}).
Among them, Cluster and Merge are pruning methods based on the context similarity of different entries \cite{wang2022efficient, martins2022efficient}.

We report case-sensitive detokenized BLEU \cite{papineni2002bleu} calculated by \textit{sacrebleu}\footnote{\url{https://github.com/mjpost/sacrebleu}} and COMET \cite{rei2020comet} calculated by publicly available \textit{wmt20-comet-da}\footnote{\url{https://github.com/Unbabel/COMET}} model.
For the prunign methods, statistical significance test \cite{koehn2004statistical} against the full datastore (Adaptive \knn) are conducted as well.



\section{Experiment Results and Analysis}
\label{sec:results}

\subsection{Safely Pruning with \method}
\label{sec:safe}

Experiment results on OPUS domains are presented in Table \ref{tab:opus_test}.
For the reference, the pre-trained NMT model usually does not translate well on target domains. 
Finetuning and Adaptive \knn have comparable performances.

We perform datastore pruning with \method for different domains and report the largest pruning ratio without significant performance degradation on the test set.

Compared with using full datastore (Adaptive $k$NN), our method (\textbf{\method}) cutting off 25\%-45\%\footnote{The best pruning ratios are different because for different target domains, the translation knowledge inside the pre-trained NMT model is different. For the target domain which is more distant away, the best pruning ratio is naturally smaller.} entries of the datastore while achieving comparable performance.
On the two largest domains, ``OPUS-Medical'' and ``OPUS-Law'', our method successfully prunes 45\% datastore (millions of key-value pairs).
Excellent pruning performance validates our analysis concerning with local correctness.

Cluster and Merge lead to a larger degradation of translation performance\footnote{Here we show the pruning effect of the two methods under a large pruning ratio. According to the original published results, these methods may suffer less performance degeneration when the pruning ratio is lower, e.g. 10\% \cite{wang2022efficient}.}, showing that entries with identical target tokens indeed have different importance in assisting the NMT model.
Simply pruning all \textit{known} entries results in a significant drop of performance (All Known). 
Pruning \textit{known} entries to the same ratio as \method also lead to degradation (Known), although it outperforms Cluster and Merge.
These comparisons indicates that the entry correctness only partially reflects entry importance, demonstrating the necessity of the neighborhood correctness analysis with knowledge margin.

The results on UM domains are presented in Table \ref{tab:um_test}.
The datastore could be pruned by 30\% for ``UM-Law'' and 15\% for ``UM-Thesis'' Datastore without any sacrifice in translation performance. 
The other findings are similar with those in German-English experiments.

\begin{table}[tbp]
    \footnotesize
    \centering
    \newcolumntype{C}[1]{>{\centering\arraybackslash}p{#1}}
    \scalebox{0.8}{
    \begin{tabular}{l|p{0.5cm}<{\centering}p{0.7cm}<{\centering}p{1.2cm}<{\centering}|p{0.5cm}<{\centering}p{0.7cm}<{\centering}p{1.2cm}<{\centering}}
    \toprule
               & \multicolumn{3}{c|}{UM-Law} & \multicolumn{3}{c}{UM-Thesis} \\
               & Ratio & BLEU$\uparrow$ & COMET$\uparrow$ & Ratio & BLEU$\uparrow$ & COMET$\uparrow$ \\
    \midrule
    Base         &  -    & 30.36   & 0.3857   & -     & 13.13   & -0.0442  \\
    Finetune          &  -    & 58.82   & 0.6375   & -     & 16.86   & -0.0295  \\
    Adaptive $k$NN    &  0\%  & 58.64   & 0.6017   & 0\%   & 17.49   & -0.0146 \\
    \midrule
    \textbf{Random}         & 30\%  & $\textrm{53.78}^*$   & $\hspace{0.12cm}\textrm{0.5661}^*$   & 15\%  & $\textrm{16.14}^*$   & $\hspace{0.12cm}\textrm{-0.0280}^*$  \\
    \textbf{Cluster}        & 30\%  & $\textrm{49.65}^*$   & $\hspace{0.12cm}\textrm{0.5274}^*$   & 15\%  & $\textrm{15.73}^*$   & $\hspace{0.12cm}\textrm{-0.0419}^*$  \\                     
    \textbf{Merge}          & 30\%  & $\textrm{56.51}^*$   & $\hspace{0.12cm}\textrm{0.5873}^*$   & 15\%  & $\textrm{17.00}^*$   & $\hspace{0.12cm}\textrm{-0.0296}^*$  \\                     
    \textbf{Known}          & 30\%  & $\textrm{56.92}^*$   & $\hspace{0.12cm}\textrm{0.5762}^*$   & 15\%  & 17.25                & -0.0143  \\
    \textbf{All Known}      & 63\%  & $\textrm{46.45}^*$   & $\hspace{0.12cm}\textrm{0.4720}^*$   & 47\%  & $\textrm{15.33}^*$   & $\hspace{0.12cm}\textrm{-0.0525}^*$  \\
    \midrule
    \textbf{\method} (ours) & 30\%  & 58.65                & 0.6056                & 15\%  & 17.52                & -0.0122  \\
    \bottomrule
    \end{tabular}
    }
    \caption{Pruning Effect on two UM domains. ``Ratio'' denotes the pruning ratio. Higher ``BLEU'' and ``COMET'' scores indicate better translation quality. ``*'' means that performance decline is statistically significant ($p<0.05$).}
    \label{tab:um_test}
\end{table}

\subsection{How Knowledge Margin Affects Pruning Performance?}
\begin{figure}[tbp]
    \centering
    \includegraphics[width=0.43\textwidth]{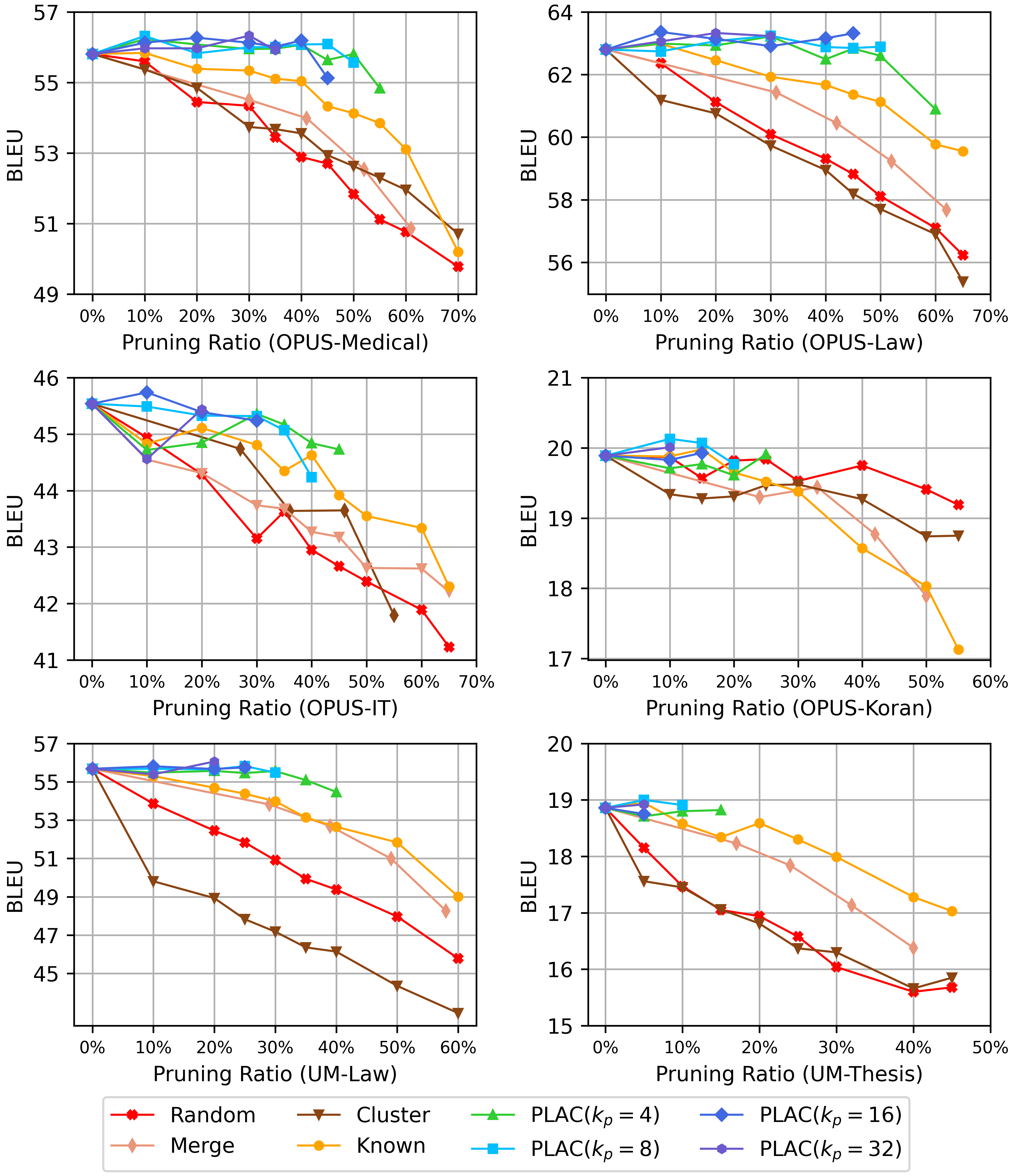}
    \caption{BLEU scores of adaptive $k$NN-MT models with pruned datastore on different domains' development set. Different symbols represent different ways of pruning datastore.}
    \label{fig:opus_dev}
\end{figure}

In this section, we examine how knowledge margin affects pruning performance and provide more insight into our proposed method.
Figure \ref{fig:opus_dev} plots BLEU scores of adaptive $k$NN-MT models with pruned datastore under different pruning ratios on development sets.
We can observe that trends are mostly similar in different domains. 
Pruning by \method achieves the best performance over the other baselines and the performance is more stable even with a higher pruning ratio. 

Note that Known is a case where neighborhood correctness is dismissed during entry pruning.
Although it outperforms Random, Cluster and Merge in most scenarios, its performance is still unstable.

When tuning the hyperparameter $k_p$ among \{4, 8, 16, 32\}, we can see a trade-off between BLEU score and the pruning ratio.
Large $k_p$ leads to a small sacrifice of BLEU score but a lower pruning ratio.
Small $k_p$ allows us to prune more entries but causes significant BLEU scores decline after a specific threshold ratio.
For example, when $k_p=4$, it is allowed to prune 55\% ``OPUS-Medical'' datastore, but translation performance declines drastically after the pruning ratio reaches 50\%.
Finally, we choose the top-right point\footnote{Hyper-parameter values of these points are reported in Appendix \ref{apx:impl}.} in each subfigure as the best-performed setting for each domain, which are used in other experiments.



\subsection{Datastore Entries With Lower Knowledge Margin Are Indeed Valuable}
In this section, we want to verify that entries with low knowledge margin are truly important for NMT adaptation.
For this purpose, we remove entries from datastore with a reversed strategy, i.e. the knowledge margin of $(h, y)$ is less than $k_p$. 

Table \ref{tab:wrong} shows pruning effect.
We can see that pruning entries with reverse strategy suffers significant performance decline at even a small pruning ratio, demonstrating the importance of these entries for domain adaptation.
We also show some cases for each domain in Table \ref{tab:corpus_case}.
We can see that target tokens of these valuable entries are more domain-specific, e.g. ``dose'' and ``Executive''.

\begin{table}[htbp]
    \footnotesize
    \centering
    \scalebox{0.9}{
    \begin{tabular}{c|ccccc}
    \toprule        
    OPUS-Law          & 10\%    & 20\% & 30\% & 40\% & 45\% \\
    \midrule

    reverse pruning & -1.91   & -4.00   & -6.19  &  -8.71  & -10.38   \\
    \textbf{\method} (ours)     & +0.00   & -0.19  &  +0.18  &  -0.21  & -0.31   \\
    \bottomrule
    \end{tabular}
    }
    \caption{Translation performance difference (BLEU) compared with Adaptive \knn using full datastore under different pruning ratios.}
    \label{tab:wrong}
\end{table}

\begin{table*}[ht]
    \footnotesize
    \centering
    \scalebox{0.9}{
        \begin{tabular}{p{2cm}<{\centering}|p{6cm}|p{8cm}}
        \toprule
        \textbf{Domain} & \textbf{Source Sentence ($\mathbf{x}$)} & \textbf{Target Sentence ($\mathbf{y}$)} \\
        \midrule
        OPUS-Medical & Die H\"ochst\atat do\atat sis sollte 30 mg - Tag nich \"uberschrei\atat ten . & The \underline{maximum dose} should not exce\atat ed 30 mg \underline{/} day .\\
        \midrule 
        OPUS-Law &  Das Direkt\atat ori\atat um entscheidet \"uber die Organisation seiner Sitz\atat ungen . & \underline{The Executive} Board \underline{shall decide on} the \underline{organisation} of its meetings . \\
        \midrule
        OPUS-IT & Sie haben eventuell einen Programm\atat fehler entdeckt . & \underline{You may have encounter\atat} ed a \underline{bu\atat} g \underline{in the program .} \\
        \midrule
        OPUS-Koran & Das ist eine schmerz\atat hafte P\atat ein . & Target sentence: \underline{That would be a grie\atat} v\atat ous \underline{aff\atat} li\atat ction \underline{.} \\
        \midrule
        UM-Law & \kt{保险公司\ 依法\  接受\  监督\  检查\ 。} & \underline{Any} insurance \underline{company shall accept supervision} and \underline{inspection according} to law . \\
        \midrule
        UM-Thesis & \kt{中国\  能源需求\  及其\  风险管理\  研究} & \underline{The Research on Energy Demand and Its Risk} Management in China \\
        \bottomrule
        \end{tabular}
    }
    \caption{Case study for remaining knowledge in different domain's pruned datastore. The underlined parts are target tokens of entries with small margin values.}
    \label{tab:corpus_case}
\end{table*}

\begin{table*}[htbp]
    \footnotesize
    \centering
    \scalebox{0.9}{
    \begin{tabular}{l|p{0.8cm}<{\centering}p{0.8cm}<{\centering}|p{0.8cm}<{\centering}p{0.8cm}<{\centering}|p{0.8cm}<{\centering}p{0.8cm}<{\centering}|p{0.8cm}<{\centering}p{0.8cm}<{\centering}|p{0.8cm}<{\centering}p{0.8cm}<{\centering}|p{0.8cm}<{\centering}p{0.8cm}<{\centering}}
    \toprule
               & \multicolumn{2}{c|}{OPUS-Medical} & \multicolumn{2}{c|}{OPUS-Law} & \multicolumn{2}{c|}{OPUS-IT} & \multicolumn{2}{c|}{OPUS-Koran} & \multicolumn{2}{c|}{UM-Law} & \multicolumn{2}{c}{UM-Thesis} \\
               &  Space & $\Delta$ &  Space & $\Delta$ & Space & $\Delta$ & Space & $\Delta$ & Space & $\Delta$ & Space & $\Delta$ \\
    \midrule
    Full Datastore & 492 & -  &  1,328  &  - & 265 & -  & 54  & - &  680 & - & 810 &  - \\
    \textbf{\method} (ours) & 279 &  43\%  &  739    &  44\%  & 166 &  37\%  & 45  & 17\%  &  479 & 30\%  & 690 & 15\%   \\
    \bottomrule
    \end{tabular}
    }
    \caption{Memory Space (MB) comparsion between pruned datastore and full datastore. ``Space'' denotes the memory space taken by the index file and ``$\Delta$'' denotes the percentage of space saved by our method.}
    \label{tab:disk}
\end{table*}

\subsection{\method Is Applicable to Different $k$NN-MT Variants}
\label{sec:robustness}
For more comprehensive evaluation, we plug our pruned datastore into different $k$NN-MT variants, i.e. vanilla $k$NN \cite{khandelwal2021nearest}, KSTER \cite{jiang2021learning} and adaptive $k$NN. 
Experiment results on OPUS-Law domain show that our pruned datastore does almost no harm to the translation performance of different variants, demonstrating the effectiveness of \method.

\begin{table}[htbp]
    \footnotesize
    \centering
    \scalebox{0.9}{
    \begin{tabular}{c|cc}
    \toprule        
     $k$NN-MT Variants       & Full  & Pruned \\
    \midrule
    Vanilla $k$NN \cite{khandelwal2021nearest}   & 61.34         & 61.24      \\
    KSTER \cite{jiang2021learning}           & 62.45         & 62.30      \\ 
    Adaptive $k$NN \cite{zheng2021adaptive}  & 63.53         & 63.22      \\
    \bottomrule
    \end{tabular}
    }
    \caption{Translation performance (BLEU) of different $k$NN-MT variants with full and pruned datastore on OPUS-Law domain's test set.}
    \label{tab:variant}
\end{table}

\subsection{Pruned Datastore Occupies Less Memory Space}
In practice, the datastore must be loaded to CPU and GPU memory during inference. So its size affects the efficiency.
Since Faiss index is used to index and represent the datastore, we compare the size of index file before and after pruning (Table \ref{tab:disk}).
For all the domains, our pruning method \method significantly reduces the memory occupation.
The ratio of saved memory space is roughly identical with the PLAC pruning ratio.
For the largest datastore, ``OPUS-Law'', the memory space can be reduced by 44\%. 


\section{Related Work}
Less attention have been paid to the research of interpretability of $k$NN-MT. 
To the best of our knowledge, we are the first to systematically study the relationship between the NMT model and the datastore.
As for datastore pruning, \citet{wang2022efficient} and \citet{martins2022efficient} prune the datastore based on the hypothesise that entries with similar translation is redundant.
Actually, entries with similar translations may have different importance to the translation. 
Our analysis suggests one way to understand these differences.


\section{Conclusion}
It is interesting to explore how a neural model and a symbolic model works together. In this paper, we propose to analyze the local correctness of the neural model's predictions to identify the conditions where the neural model may fail.
By introducing a knowledge margin metric to measure the local correctness, we find that the NMT model often fails when the knowledge margin is small.
These results provide support for building a more explainable machine translation system.

Based on analyses, we can safely prune the datastore with the proposed \method method.
Empirically, the datastore could be successfully pruned up to 45\% while retaining translation performance. 
This results validate our earlier findings about the local correctness and translation failures.

Our method is general to different \knnmt variants and easy to implement.
Future directions maybe using local correctness to explore more interpretability issue of NMT domain adaptation, e.g. catastrophic forgetting.

\section{Ethical Considerations}
In $k$NN-MT works, the symbolic datastore helps adaptation but also introduce privacy concerns.
Since $k$NN-MT explicitly saves all target language tokens in the datastore, there is a risk of privacy leakage.
In the future, more efforts may be put into addressing this issue.

\normalem
\bibliographystyle{acl_natbib}
\bibliography{acl}

\clearpage
\appendix

\section{Detailed descriptions of Target Domains}
\label{sec:description}
``OPUS-Medical'' domain is made out of PDF documents from the European Medicines Agency.
``OPUS-Law'' domain is a collection of the legislative text of the European Union.
``OPUS-IT'' domain is constructed from localization files and documents of GNOME, KDE, PHP, Ubuntu, and OpenOffice.
``OPUS-Koran'' domain is a collection of Quran translations complied by the Tanzil project.
``UM-Law'' domain contains law statements from mainland China, Hong Kong, and Macau.
``UM-Thesis'' domain is composed of journal topics in the research area, including electronics, agriculture, biology, economy, etc.

\section{Involved Scientific Artifacts}
In this section, we list the artifact used in our project:

\textit{Moses (LGPL-2.1-License)}: It is a statistical machine translation system that allows you to automatically train translation models for any language pair.

\textit{Jieba (MIT-License)}: it is a library for chinese word segmentation.

\textit{Subword-nmt (MIT-License)}: Subword-nmt is a package containing preprocessing scripts to segment text into subword units.

\textit{Fairseq (MIT-license)}: It is a sequence modeling toolkit that allows researchers and developers to train custom models for translation, summarization, language modeling and other text generation tasks.

\textit{Faiss (MIT-license)}: It is a library for efficient similarity search and clustering of dense vectors.

For the sake of ethic, our use of these artifacts is consistent with their intended use.

\section{Implementation Details}
\label{apx:impl}
We implement adaptive $k$NN-MT with \citet{zheng2021adaptive}'s released code and script\footnote{\url{https://github.com/zhengxxn/adaptive-knn-mt}} based on \textit{fairseq}\footnote{\url{https://github.com/pytorch/fairseq}} \cite{ott2019fairseq}.
Due to the large space of hyper-parameters, we follow \citet{zheng2021adaptive} to set the number of retrieved entries ($k_a$) as 8 when training adaptive $k$NN-MT models for most experiments, and report pruning performance under different $k_a$ in Appendix \ref{sec:hyper}.
During inference, we set beam size as 5 and length penalty as 1.0.

For implementing \method, the hyper-parameter $k_p$ in Algorithm \ref{alg:pruning} implicitly determines the maximum number of entries that are allowed to be pruned. So we tune  $k_p$ among the subset of \{4, 8, 16, 32\} when given different pruning ratio $r$.

After buiding the datastore, we follow previous $k$NN-MT works \cite{khandelwal2021nearest, zheng2021adaptive} and use Faiss\footnote{\url{https://github.com/facebookresearch/faiss}} index \cite{johnson2019billion} to represent the datastore and accelerate nearest neighbors search.

In Table \ref{tab:hyperparameter}, we report hyperparameters to reproduce our main results in Table \ref{tab:opus_test} and \ref{tab:um_test}.
In our experiments, it takes at most 1.5 GPU hours to train adaptive $k$NN-MT models on a single NVIDIA Titan RTX.

\begin{table}[htbp]
    \footnotesize
    \centering
    \scalebox{0.9}{
        \begin{tabular}{l|cccc}
        \toprule
        Target Domain & $k_p$ &  $r$  &  $k_a$  & $T$ \\
        \midrule 
        OPUS-Medical &  8   & 45\%   & 8   & 10  \\
        OPUS-Law     &  16  & 45\%   & 8   & 10  \\
        OPUS-IT      &  4   & 40\%   & 8   & 10  \\
        OPUS-Koran   &  4   & 25\%   & 8   & 100 \\
        UM-Law       &  4   & 30\%    & 8   & 100 \\
        UM-Thesis    &  4   & 15\%    & 8   & 100 \\
        \bottomrule
        \end{tabular}
    }
    \caption{Hyperparameters for pruning datastore and training adaptive $k$NN-MT models.}
    \label{tab:hyperparameter}
\end{table}

\section{Pruning Effect is Insensitive to Hyperparameter $k_a$}
\label{sec:hyper}
To demonstrate the reliability of our pruned datastore, after pruning datastore, we train adaptive $k$NN-MT models with different hyperparameter $k_a$ and evaluate their translation performance (BLEU) on ``OPUS-Law'' domain's test set (Table \ref{tab:ka}). 
Results show that our pruning method enjoys consistent performance under different $k_a$.

\begin{table}[htbp]
    \footnotesize
    \centering
    \scalebox{0.9}{
    \begin{tabular}{c|cccc}
    \toprule        
     OPUS-Law       & $k_a=4$ & $k_a=8$ & $k_a=16$ & $k_a=32$ \\
    \midrule
    Adaptive $k$NN  &  63.31   & 63.53    & 63.56     & 63.33\\
    \textbf{\method} (ours)  &  62.93   & 63.22    & 63.18     & 63.22 \\
    \bottomrule
    \end{tabular}
    }
    \caption{Pruning performance under different $k_a$.}
    \label{tab:ka}
\end{table}
\end{document}